\documentclass{article}
\usepackage{cite}
\usepackage{amsmath,amssymb,amsfonts}
\usepackage{algorithmic}
\usepackage{graphicx}
\usepackage{textcomp}
\usepackage{xcolor}
\usepackage{authblk}
\usepackage[caption=false]{subfig}
\def\BibTeX{{\rm B\kern-.05em{\sc i\kern-.025em b}\kern-.08em
    T\kern-.1667em\lower.7ex\hbox{E}\kern-.125emX}}
\usepackage{array}
\newcommand{\norm}[1]{\left\lVert#1\right\rVert}
\newenvironment{ColorPar}[1]{%
    \leavevmode\color{#1}\ignorespaces%
}{%
}%

\begin{document}

\title{eBIM-GNN : Fast and Scalable energy analysis through BIMs and Graph Neural Networks\\
\thanks{Funding Agency: National Geospatial Programme, Department of Science and Technology, Government of India}
}


\author{
  Rucha Bhalchandra Joshi\\
  \texttt{rucha.joshi@niser.ac.in}
  \and
  Annada Prasad Behera\\
  \texttt{annada.behera@niser.ac.in}
  \and
  Subhankar Mishra\\
  \texttt{smishra@niser.ac.in}
}
\affil{
    National Institute of Science Education and Research,
    Bhubaneswar, Odisha 752050\\
    Homi Bhabha National Institute, 
    Anushaktinagar, Mumbai 400094\\
}


\date{}


\maketitle

\begin{abstract}
Building Information Modeling has been used to analyze as well as increase the energy efficiency of the buildings. It has shown significant promise in existing buildings by deconstruction and retrofitting. Current cities which were built without the knowledge of energy savings are now demanding better ways to become smart in energy utilization. However, the existing methods of generating BIMs work on building basis. Hence they are slow and expensive when we scale to a larger community or even entire towns or cities. In this paper, we propose a method to creation of prototype buildings that enable us to match and generate statistics very efficiently. Our method suggests better energy efficient prototypes for the existing buildings. The existing buildings are identified and located in the 3D point cloud. We perform experiments on synthetic dataset to demonstrate the working of our approach. 
\end{abstract}


\section{Introduction}
The Building Information Model (BIM)\cite{azhar11bim} is an instance of a populated data model of buildings. It constitutes of all the different kinds of data pertaining to a specific building. These data invariably or uniquely define the same building. It is not a temporal representation; however a static projection of the building that defines the building from various different perspectives across disciplines. BIM has replaced traditional onsite construction process in being a significant workload percentage of the AEC industry \cite{darko2020building}. 

In the recent years, BIM has been extended to include specialised tools such as energy analysis, structural analysis for energy and space management \cite{volk2014building}. However, creation of accurate BIMs is time consuming and expensive \cite{ahmed2018barriers} if it is done for the existing buildings and the advantages it offers for energy efficiency calculations \cite{eleftheriadis2017life, ghaffarianhoseini2017application} are lost. Hence it is imperative that we look for scalable and low cost approximate BIM modeling for existing infrastructure in a city or a local community. 

In this paper, first we create the notion of the prototype buildings which serve as the state of the art implementation of energy efficiency techniques given particular conditions of space, number of residents and energy consumption are considered. Secondly, 
we give a method to suggest a better prototype alternatives for the existing building. We propose using Point-GNN for detecting the buildings and determining their bounding boxes and then suggesting the efficient prototypes for the detected buildings.

Contributions of the paper are as follows:
\begin{itemize}
    \item Introducing the novel prototype buildings for capturing the energy profiles. 
    \item Scalable framework for energy efficiency analysis and recommendation.
    \item Use Point-GNN for identifying existing buildings and suggesting energy efficient prototypes for them 
    \item We perform the experiments on completely synthetic data for demonstrating our proposed approach
\end{itemize}

The remainder of this paper is organized as follows. We review some of the important related studies in section \ref{sec:related}. We discuss the prototype building information model (BIM) in detail in section \ref{sec:prototypeBIM}. We then discuss out proposed method in section \ref{sec:methodology}. We then conclude with section \ref{sec:conclusion} and discuss the future works.

\section{Related Works}
\label{sec:related}
Climate change and impending global energy crisis has moved the scientists along with Architecture, Engineering and Construction (AEC) industry have called for wide use of BIM-based energy analysis for building performance assessment \cite{chen2017green}. Although creating BIMs for exisiting building and retrofitting them to improve the existing building management and maintenance has been done in practice, it is important to note a few caveats. In \cite{di2015bim}, the researchers have tried to update a school building. Even though results are significant, so is the time as well as expense of creating an accurate BIM. 

Similar exercises were also carried out by the authors in \cite{bruno2018historic} for historic buildings where they added the diagnosis part for automating acquisition and update of knowledge base for heritage sites. Deconstruction of existing buildings were also taken up by the research community by building 3D models through devices such as kinect for generating low accurate and optimized 3D BIMs in real time \cite{volk2018deconstruction}. However, the problem of scalability still remained an open challenge as none of the methods could give us a quick answer for a larger community such as a city to go through the tedious and time consuming process of the creating individual BIMs for buildings.

There are three broad approaches that are taken to handle the tasks related to point clouds using Deep Neural Networks. The first approach takes considers the point cloud as a set of points. The Multi Layered Perceptron (MLP) is used to extract the features corresponding to each of the nodes. \cite{zaheer2017deep} and \cite{qi2017pointnet} are such networks that aggregate the feature vectors obtained by using neural networks for the set of points.

The other approach considers the point cloud as a grid. The points in the 3D grids units called voxels are considered in \cite{zhou2018voxelnet} and applies 3D convolutions onto it for object detection. Some approaches \cite{chen2017multi}, \cite{yang2018pixor} project the point cloud into 2D space to apply the 2D convolutions onto it. However, projections into grids has drawback of regularity that is added due to the regular nature of the grid, unlike the point cloud.

The third approach is to model the point cloud using an irregular structure of graphs. The graph neural networks in its every layer updates the vertex/edge features by aggregating them across their neighborhood\cite{wu2020comprehensive}, \cite{Joshi2022}. Unlike 2D/3D convolutions, graph convolutions does not reduce the number of points in every iterations. \cite{Point-GNN} first systematically converts the point cloud into a graph for object detection and also gives the mechanism for determining the bounding box of the detected objects.

\section{Prototype Building Information Model (BIM)}
\label{sec:prototypeBIM}

\subsection{3D Model}
\begin{figure}[h]
    \centering
    \includegraphics[width=\linewidth]{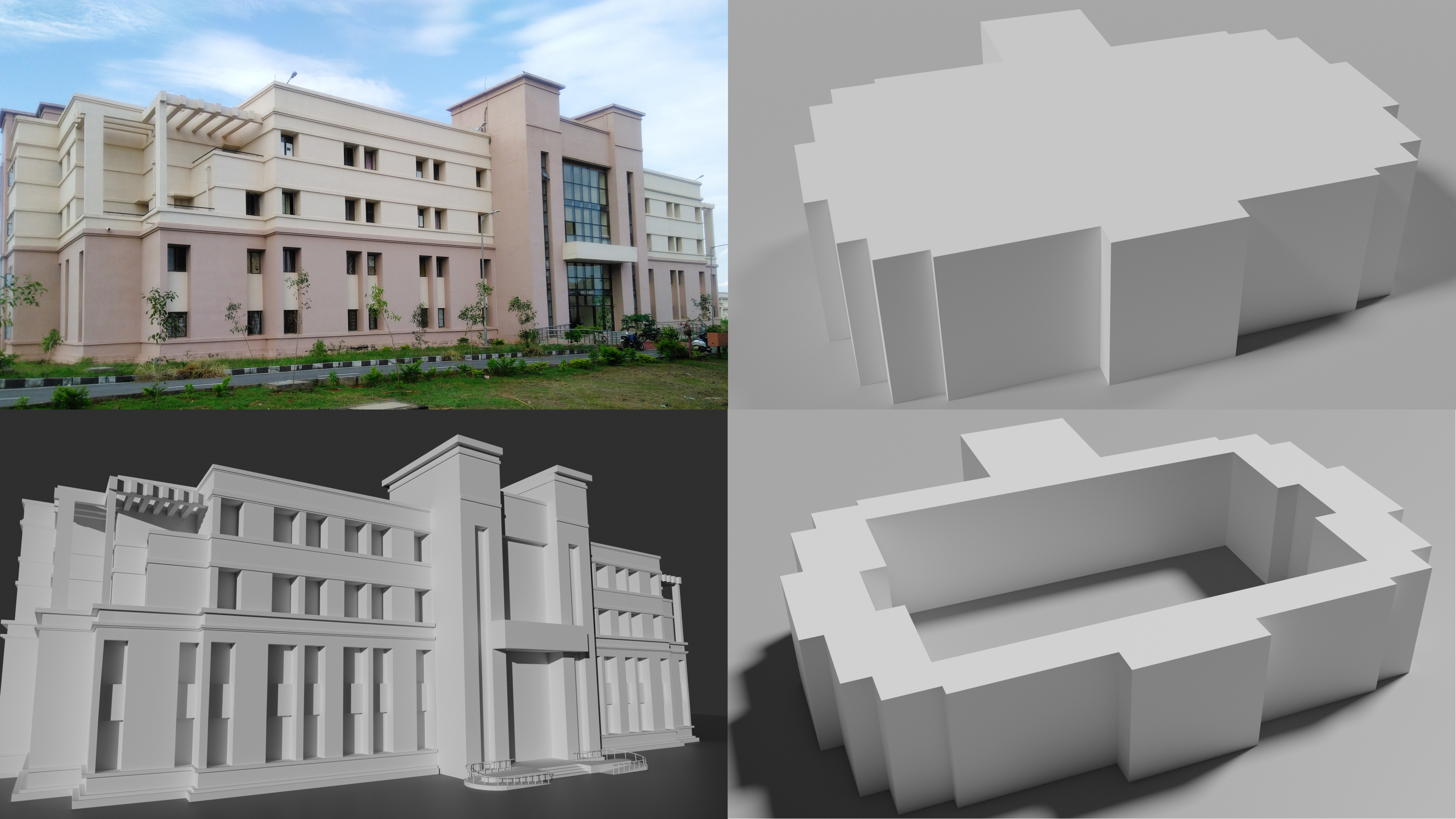}
    \caption{3D view of different level of details. (Top left) Original photo. (Bottom left) High level of detail. (Top right) Low level of detail. (Bottom right) Medium level of detail.}
    \label{fig:lod}
\end{figure}

3D model gives a visual feedback for the geometry the structure\cite{baz04vbe}. The BIM model is abstract which contains only information but it is not easy to visualize BIM. Since the BIM also contains information about the geometry of the building, they are easily accurately rendered to produce a visual information BIM. In fact, the 3D model is used to accurately find the geometry of the various structure in BIM, and not the other way around, since, obtaining the geometry is difficult humanly without a visual feedback. 3D model with various levels of details are shown in Figure \ref{fig:lod}.


\subsection{Data collection}
The first step for the creation of BIM for an entire city requires careful planning and collection of data. The BIM in itself is a very data intensive model. A assortment of BIMs for each building in a city increases the amount of data collected by many folds. Some of which may be redundant while many are not. But even without taking into factor the huge amount of data to be collected, there are many other factors that come to play.

First, it is equally challenging to figure out which data can be provided by whom and at what time. Even though a overall summary of the data maybe made available, the exact data points are not easy to be maintained and many legacy systems put in place in modern infrastructure do not even collect the data. Since, many systems both modern and legacy infrastructure is in place, there is always a risk of obtaining incomplete data and unbeknownst to the observer, the data might be scaled down and incorrect information creeps into the model.
Second, in a city, many private parties are involved who do not want to share data due to privacy and other concerns. 


With the above facts taken into consideration, the data to be collected is then divided into two general category, the geometric and non-geometric data. The geometric data is concerned mostly with the visual look of the buildings and the non geometric data with any other information that the BIM is concerned with.

The geometric data contains the GIS location, structure, space, district/area around the building. The geometric data can be used for the visualization of BIMs and as the whole, the BIM integration of the entire city. The BIM of a given building is different from that of a city, as the later contains more information than the former. Along with the above information, the BIM of the city has information, for instance, the type of area the building is located in industrial, residential or commercial. The geometric data like the roads, bridges, and other public utilities are also information that not the part any given building but of the city. This will be helpful when deciding the prototypes of the building. And the data can be used to estimate the other non-geometric data as well. 

The non-geometric data contains, information like mechanical data water requirements, electrical data like electric power consumption, HVAC data, safety or maintenance data. These non-geometric data is specific to the building's BIM, the BIMs for the entire city also needs data like the sewage, traffic, garbage (and waste disposal), drainage system etc, has to be collected. 

Another type of data that is required for creating the BIMs of the entire city, which is not required by individual building BIM is, source. For instance these source include, power like the electric grid, the water tanks that provide the water for the city and the traffic load a road can take. Such data required for creating the model for the resource constraint.

Once, all the required data that has to be collected is determined, depending on how many parameters the model has to work on, the data has to be sanitized. The data sanitization step mostly consists of checking for erroneous data, missing or incomplete data, statistically improbable data, but there are other sanitization steps specific to the type of data. The data is also to be normalized, brought to reasonable range with respect to both time and space, convert all the different data units to one uniform unit.

The data collection is a continuous process. As the city grows, the new buildings are created and old buildings are either demolished or renovated, the model evolves with time as the new data enters the system. Optionally, the model should be able to make prediction about the BIM instead of only querying or suggest future development and planning. This adds new data to the system.

\subsection{Prototype Buildings}
There are many buildings in a city and each building is different from the rest. It is true that no two building are similar in all the features of data collected. BIMs generally deal with individual buildings and therefore, making a detailed model is not an issue while, for a city with creating BIMs for all the building is not only a huge task, it doesn't provide any additional benefit to the model. Any perceived advantage is further diminished, since it becomes a hassle and un-maintainable on a large scale, while not providing any significant advantage over a statistical measure.

A alternative approach is, based on the data, to create a different prototypes of buildings and use the prototypes of building to replace all the buildings with their respective BIMs. The prototypes maybe detailed with respect to a certain feature, while being very coarse in an different feature. A feature being coarse means that a prototype can be used to represent many buildings whose feature in question fall in a certain range. While, a detailed feature begs for an exact or near exact match of the feature.

\begin{figure}
    \centering
    \includegraphics[width=\linewidth]{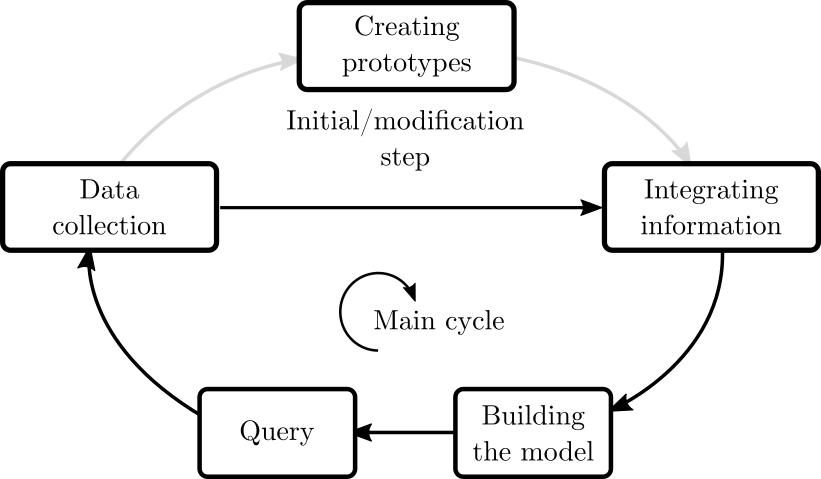}
    \caption{Life cycle of BIM for cities}
    \label{fig:cycle}
\end{figure}

Creating a prototype is not a routine part of the model. The prototype is created when the model is initially created and again when there is a necessity for it's creation. Initially, when BIMs for a city is created from the ground up, a bulk of the work goes into the creation of the prototypes since many of them has to be created. But when the model has been set up, there is not a need to create prototypes anymore. But as the model evolves with more data and new prototype buildings has to be made. But creation of new prototypes is few and far in between and not a regular part of the life cycle as illustrated in the Fig. \ref{fig:cycle}. 

The creation of prototypes also depends on the coarse or the detailed feature of the data. For features that need exact or near exact match of the features of the data of the buildings with the prototype, a prototype for every such match is required. Therefore, these features must be prominent features and must broadly divide the buildings into a small limited category. For instance, the type of the building, whether residential, commercial or industrial is such a feature, but the power consumption is not such a feature, since for every value of power consumption, there has to be a prototype, which defeats the entire purpose of making prototypes.

Power consumption, however, is coarse feature for a prototype and therefore it's prototype must span over a large range of values. The ranges of these features ideally may be chosen appropriately so that the buildings in the city may fall as evenly as possible, but this may not be possible as the number of features increases.

There are many different types of buildings at NISER. The buildings are mostly categorized based from their top down view as shown in Table \ref{tab:niserbuild} . What is most important for a building is that it should retain an overall coarse structure for any air or drone traffic. Maintaining the detail structure of the building is not only adds a lot of data points to the storage but also adds a hit to the rendering of the buildings and also for the machine learning algorithm. Making a detailed structure to the building is a ``hard'' task. But making the structure of the building too coarse, implies that the building becomes a cuboid, which is what is to be avoided. A trade-off is needed and hence a top down view of the building is sufficient enough, while maintaining the overall structure. Also, for a machine learning algorithm it is easier to create the top-down view of the buildings.

\begin{table*}
\caption{Array of 3D models and their corresponding satellite images}
\label{tab:niserbuild}
\begin{tabular}{|c|c|c|c|}
		\hline
	\includegraphics[width=.22\linewidth]{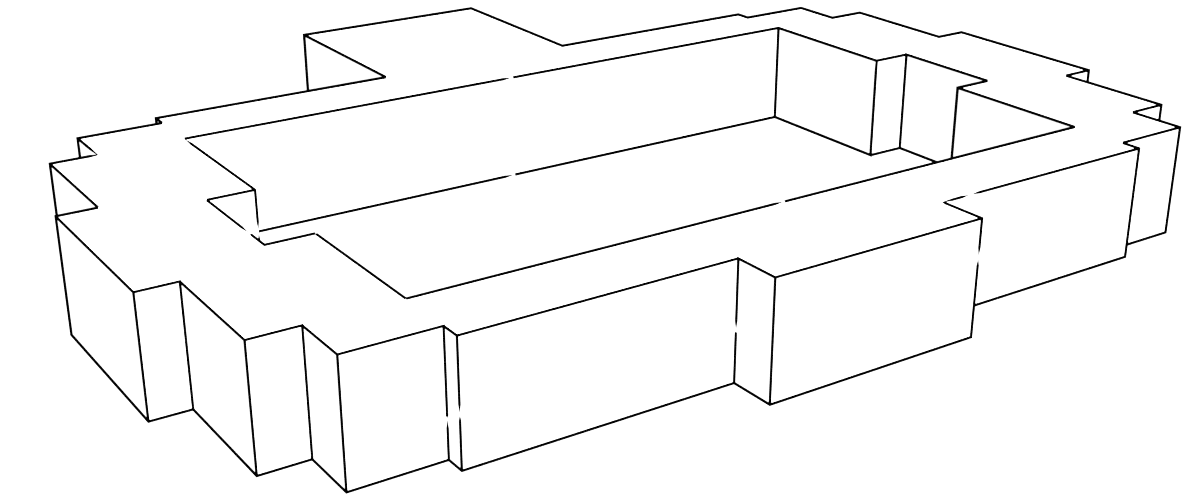}&\includegraphics[width=.22\linewidth]{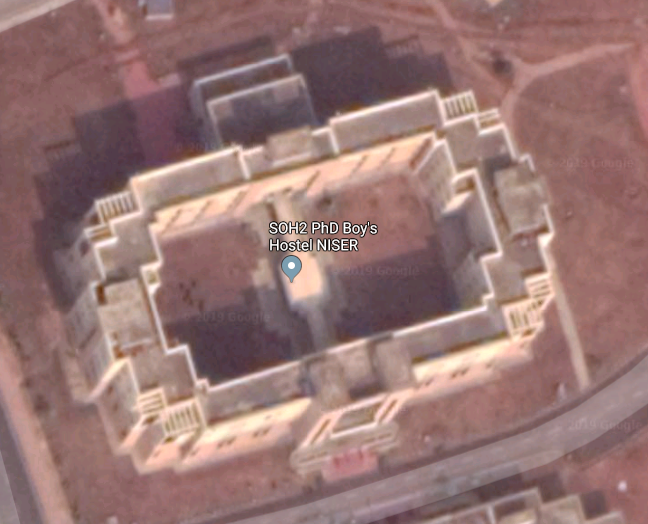} &
	\includegraphics[width=.22\linewidth]{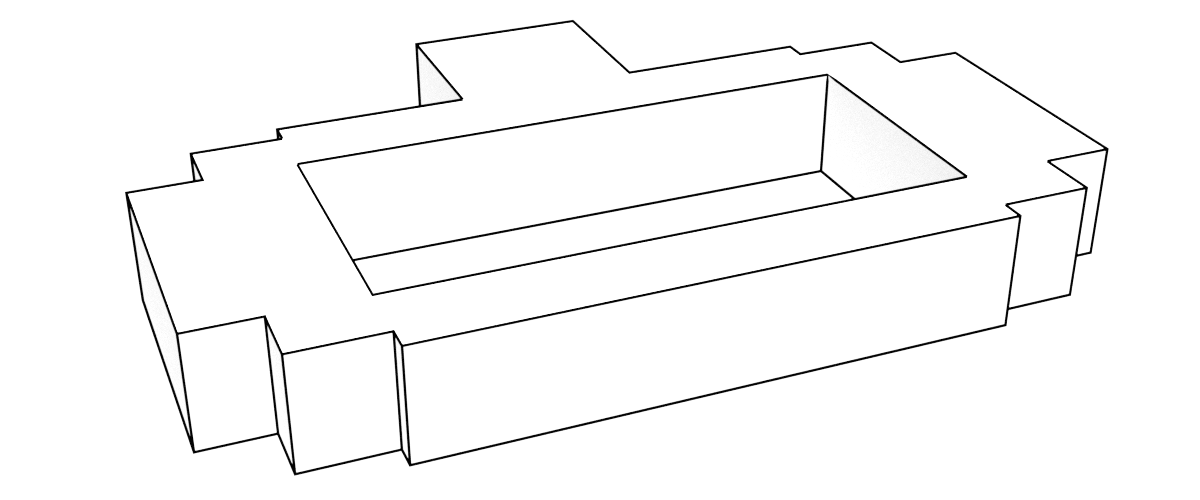}&\includegraphics[width=.22\linewidth]{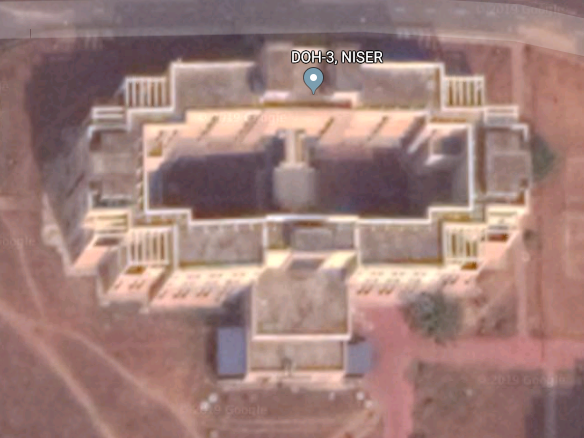}\\\hline
	\includegraphics[width=.22\linewidth]{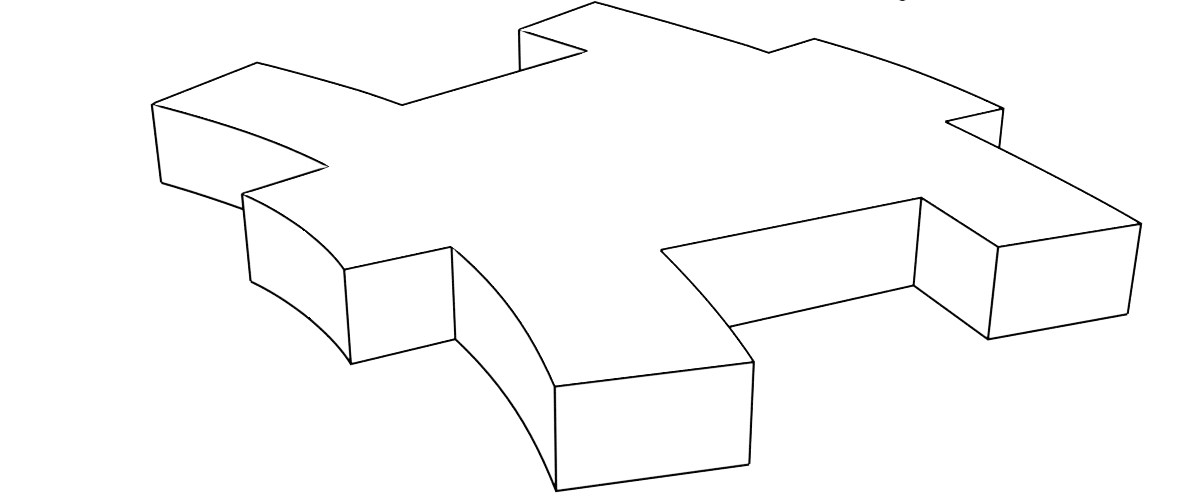}&\includegraphics[width=.22\linewidth]{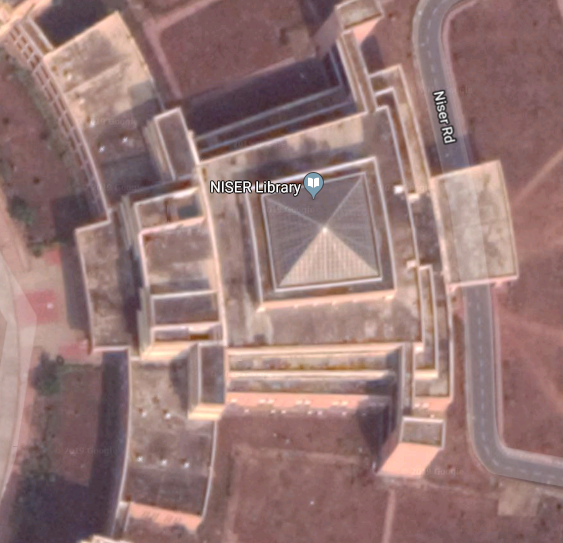} &
	\includegraphics[width=.22\linewidth]{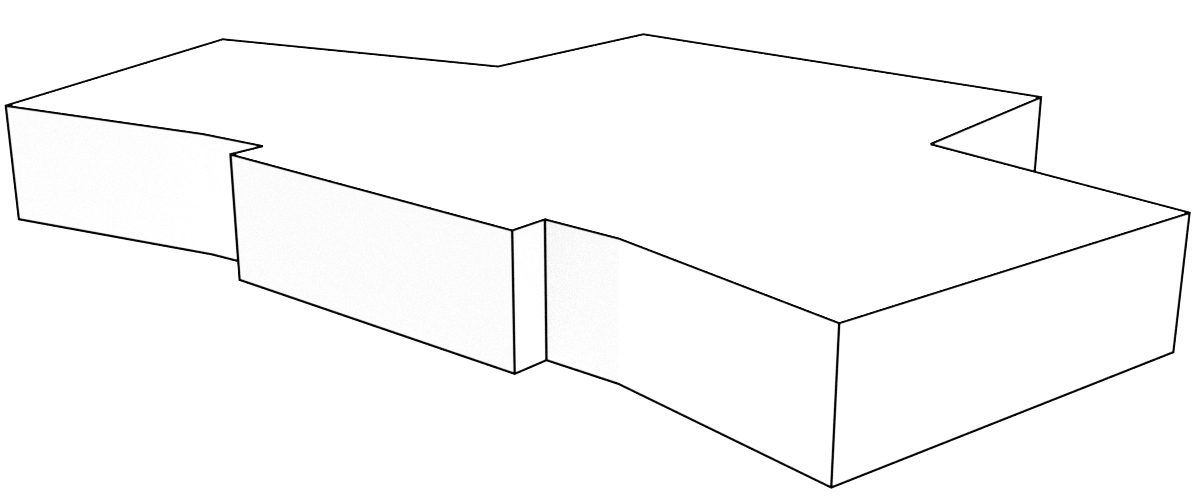}&\includegraphics[width=.22\linewidth]{./images/SoH-m.png}\\\hline 
	\includegraphics[width=.22\linewidth]{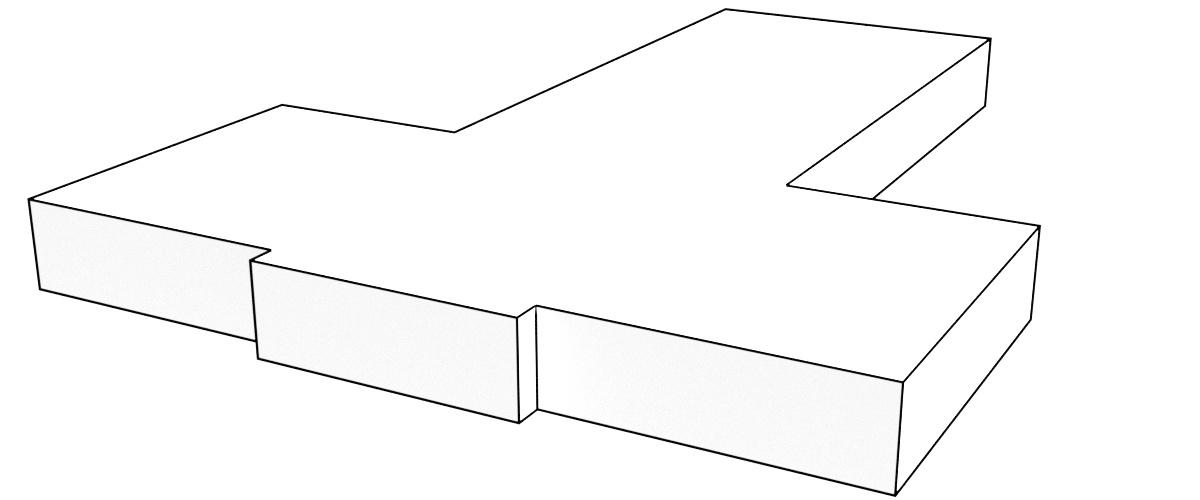}&\includegraphics[width=.22\linewidth]{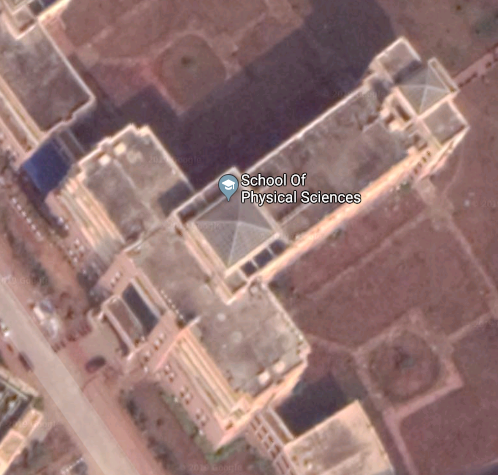} &
	\includegraphics[width=.22\linewidth]{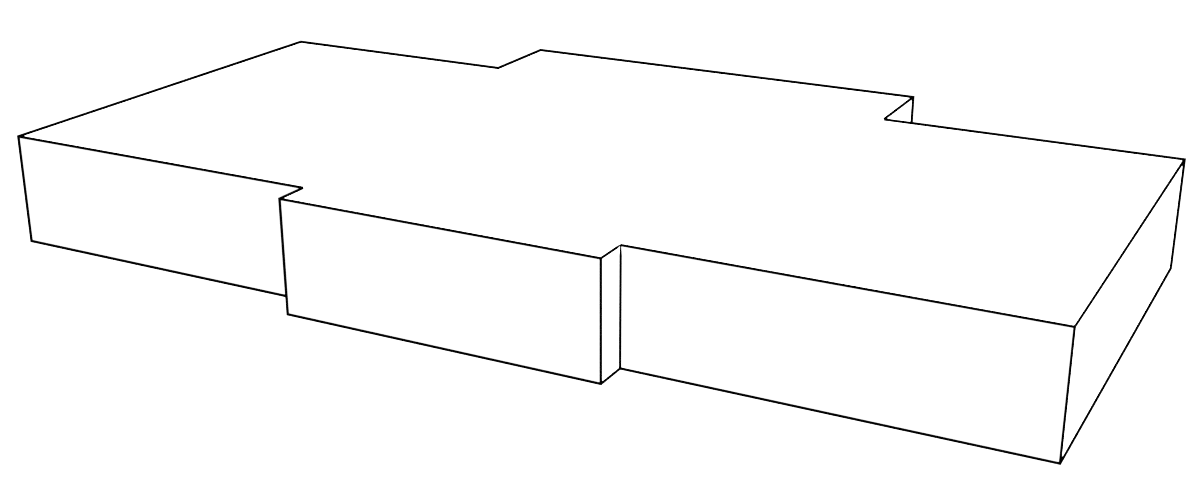}&\includegraphics[width=.22\linewidth]{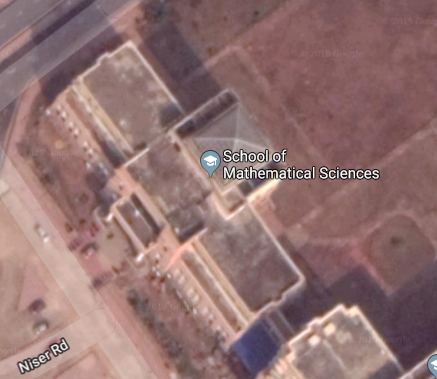}\\\hline
	\includegraphics[width=.22\linewidth]{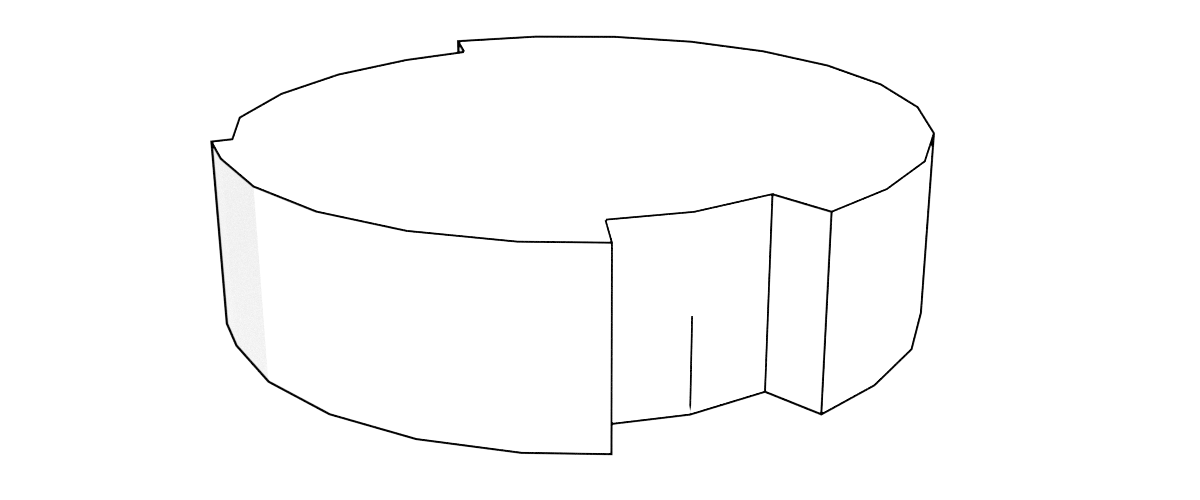}&\includegraphics[width=.22\linewidth]{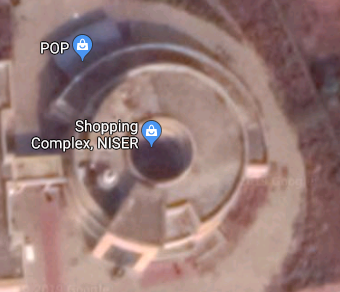} &
	\includegraphics[width=.22\linewidth]{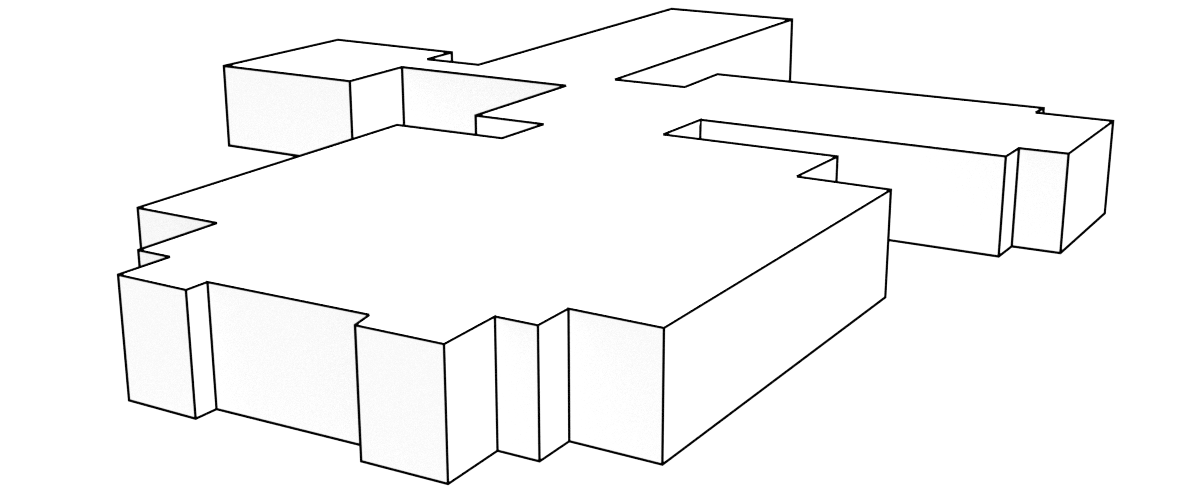}&\includegraphics[width=.22\linewidth]{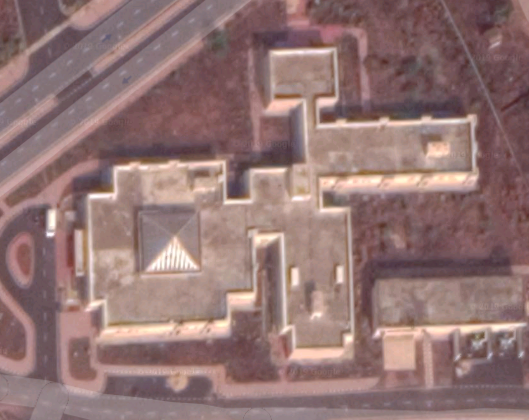}\\\hline
	\hline
\end{tabular}
\end{table*}

\begin{figure}
    \centering
    \includegraphics[width=\linewidth]{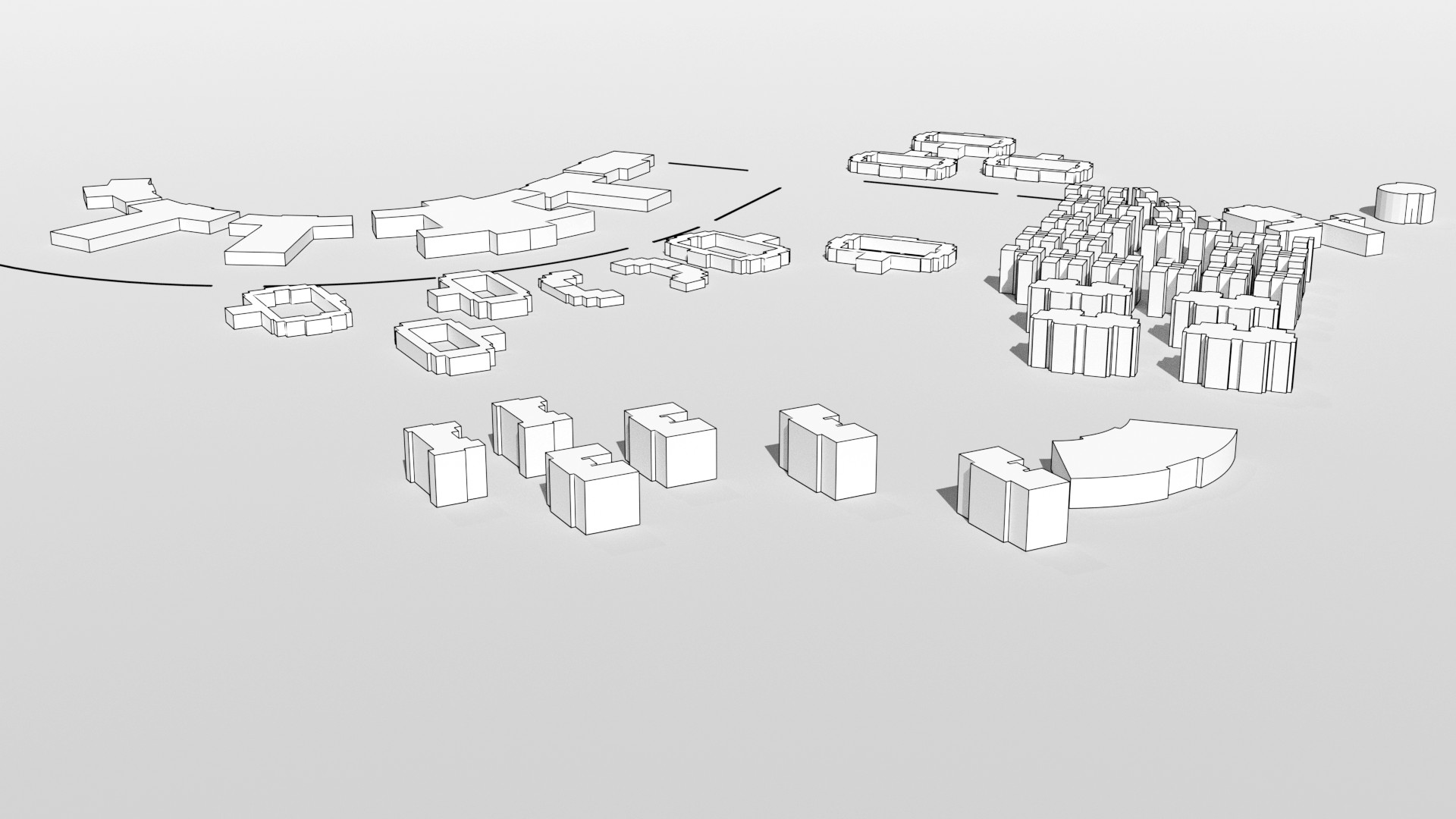}
    \caption{Entire Map of NISER}
    \label{fig:NISERcomplete}
\end{figure}

Many mapping services like Google Maps, Here Maps, OpenStreetMaps, Bing Maps etc, have top down views of the area where buildings can be easily made out. Looking at a top down view, the buildings at NISER can be categorised as such. The entire map with the protoypes at NISER for demonstration purpose is mapped in Figure \ref{fig:NISERcomplete}. 

\subsection{Storing building prototypes}

Building prototypes have to be saved in data structures. There were many candidates for the data structures that were considered. One way was to have other data in the building's geometry while the other was to incorporate the geometry in the building data structure, shown below or have a separate data structure for the buildings that includes geometry:
\begin{itemize}
    \item Building - vertices, edges, faces, power, water, traffic
    \item Building - Geometry (Vertices, edges, faces) , power, water, traffic
\end{itemize}

In the long term, the scalability of such a data structure is important. The data structure must be able to add new data categories and integrate well. Finally, keeping in mind that the ultimate goal is to have a learning algorithm to automate many of these tasks, it is imperative that the data structure should also facilitate modification automatically.

%

\section{eBIM-GNN}
\label{sec:methodology}


We are interested in a system that matches the right prototype to the buildings. Towards this goal, 
we propose a two-step system to address this problem, where the first part detects the objects in the scene and classifies into one of the two classes, building or not a building. Along with this classification, the bounding boxes of each of the detected objects are predicted, these are used to localize the buildings. The second part considers the buildings detected by the first part and further assigns the closest prototype to it.

\begin{figure}
    \centering
    \subfloat{
        \includegraphics[width=\linewidth]{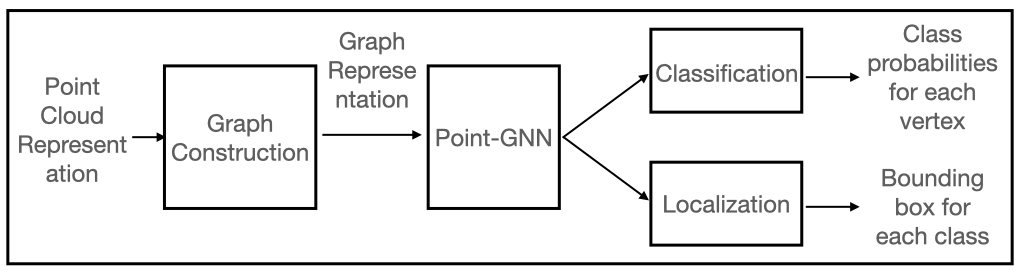}
    }
    
    \subfloat{
        \includegraphics[width=\linewidth]{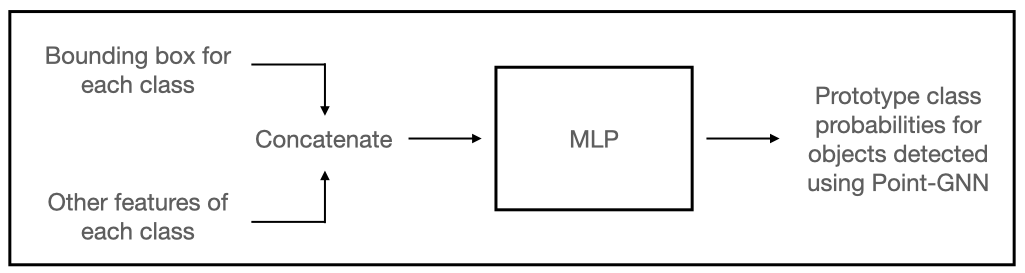}
    }
    \caption{Predicting prototypes corresponding to the detected objects}
    \label{fig:prototype-pred}
\end{figure}

\subsection{Graph Construction}
\label{subsection:prelim}
A point cloud is defined as a set $P$ of points, given as $P= \{ p_1, ... p_N \}$, where $p_i = ( x_i, s_i )$ is a point with $x_i \in \mathbb{R}^3$ are the co-ordinates of the points and $s_i \in \mathbb{R}^m $ is the feature vector of length $m$ corresponding to the point $i$. We construct a graph $G=(V, E)$ from the given point cloud. We consider each point from $P$ as the vertices in the graph. Therefore, $V = P$, and the edges in the graph $G$ we connect two nodes if they lie within a distance $r$ from each other. This is given in equation \ref{eq:edges}.

\begin{equation}
    \label{eq:edges}
    E = \{(p_i, p_j) \mid \norm{x_i-x_j}_2 \leq r \}
\end{equation}
where $r$ is the radius and $x_j$ and $x_j$ are the nodes in the graph.
We form a graph from the given point cloud as described previously. The graph formed using the entire point could might overburden the computation. Hence, before constructing the graph, the point cloud can be downsampled. Point-GNN makes use of voxel downsampling.

The initial feature vector $s_i$ of node $i$ is obtained by using $Max$ function to aggregate the embeddings generated by a multilayer perceptron (MLP) that embeds the relative coordinates and other features.

\subsection{Graph Neural Networks}
The Graph Neural Networks (GNNs) address the questions related to systems that can be modeled as graphs.
In each layer $l+1$ of a GNN, the features of the node's neighboring nodes are extracted by aggregating them, as shown in the equation \ref{eq:node_update}. 

\begin{align}
    v_i^{l+1} &= g^l (\rho(\{ e_{ij}^l \mid (i, j) \in E\}), v_i^l) \label{eq:node_update} \\
    e_{ij}^{l} &= f^t(v_i^l, v_j^l) \label{eq:edge_update} 
\end{align}
where $v^l$ and $e^l$ are feature vectors corresponding to vertex $i$ and edge $ij$ respectively. The set function $\rho$ aggregates the edge features for each vertex, function $g^l$ updates the next layer vertex features, and the function $f^l$ aggregates the vertex features to update the edge features. 

\subsection{Point-GNN}
\subsubsection{Overview}
The method Point-GNN\cite{Point-GNN} makes use of GNNs for detecting objects given a point cloud, it also localizing the bounding box of the object. Poing-GNN uses an auto-registration mechanism that enables the points to align their coordinates based on their features. The auto-registration mechanism reduced the translation variance. The alignment offset is predicted using the structural features from the previous layers in a GNN. We modify the equation \ref{eq:node_update} to update the feature vector $s_i$ corresponding to each node $i$, and add auto-registration to it.
The auto-registration mechanism is as follows:
\begin{align}
    \Delta x_i^l &= h^l (s_i^l) \\
    s_i^{l+1} &= g^l \left( \rho \left( \left\{ f^l(x_j - x_i, s_i^l) \mid (i, j) \in E \right\} \right), s_i^l \right) 
\end{align}
where $\Delta x_i^l$ is coordination offset, calculated by the function $h^l$ using the center node $i$'s feature vector $s_i^l$ from previous layer $l$. 

The functions $f^l, g^l, $ and $ h^l$ are modeled using MLPs and $\rho$ is chosen to be $Max$ function. For each layer of the Point-GNN, the MLPs are different that are not shared across the layers. As the Point-GNN is used to handle two tasks, classification and localization, the representations produced by Point-GNN after $L$ layers is passed to two branches to handle each of the two tasks.

\subsubsection{Box Merging and Scoring}
In order to improve the localization of the bounding box, Point-GNN uses the modified Non-maximum suppression(NMS). Multiple bounding boxes can be predicted as there are many points on the same object. Along with the bounding box, the corresponding confidence score is calculated. It is done using the ratio of occupied volumen, called occlusion factor. A box $b_i$ has length $l_i$, width $w_i$ and height $h_i$ and three vectors $v_i^l, v_i^w, v_i^h$ are the unit vectors in their directions respectively. Then the occlusion factor is given as
\begin{equation}
    o_i = \frac{1}{l_iw_ih_i} \prod_{v \in {v_i^l, v_i^w, v_i^h}} \max_{p_j \in b_i}(v^T x_j) - \min_{p_j \in b_i}(v^T x_j)
\end{equation}



\subsection{Loss}

The classification branch outputs a multiclass probability distribution corresponding to each of the vertex. It is given by $\{ p_{c_1}, p_{c_2}, \dots, p_{c_M} \}$ for $M$ classes of objects.
The cross-entropy loss contributed by the first branch of classification, is given as follows:
\begin{equation}
    l_{cls} = - \frac{1}{N} \sum_{i=1}^{N} \sum_{j=1}^{M} y_{c_j}^i log(p_{c_j}^i)
\end{equation}
where $p_{c}^{i}$ and $y_{c}^i$ are the predicted probabilities and one-hot encoding of the predicted label of $i$th class. 

The bounding box is predicted for every detected objects as a 7-tuple $b=(x, y, z, l, b, w, \theta)$ with 7 degrees of freedom. It consists of the coordinates of center $x, y, z$, length, width and height $l, h, w$ respectively and the yaw angle $\theta$ of the bounding box. The bounding box is encoded with vertex coordinates $(x_v, y_v, z_v)$ as follows:

\begin{align}
    \begin{split}
        \delta_x &= \frac{x-x_v}{l_m}, \delta_y = \frac{y-y_v}{h_m}, \delta_z = \frac{z-z_v}{w_m} \\
        \delta_l &= \log\frac{l}{l_m}, \delta_h = \log\frac{h}{h_m}, \delta_w = \log\frac{w}{w_m} \\
        \delta_\theta &= \frac{\theta - \theta_)}{\theta_m}
    \end{split}
\end{align}

    The loss from the localization branch that predicts the bounding box, is computed using Huber loss \cite{huber1964robust} on ground truth and prediction. The vertices that need not be localized or lie outside the bounding box contribute nothing to the loss. The average localization loss is given by:
\begin{equation}
    l_{loc} = \frac{1}{N} \sum_{i=1}^N \mathbb{I}(v_i \in b_{interest}) \sum_{\delta \in \delta_{b_i}} l_{huber}(\delta - \delta^{gt})
\end{equation}

\begin{ColorPar}{black}
\section{Prototype Matching}

We need to match the buildings with the prototypes, so that depending on the energy efficiency, a better prototype could be suggested for the given building. We propose two approaches to do this. 
\end{ColorPar}

\begin{ColorPar}{black}
\subsection{Approach 1: Parameterized Aggregation}
The bounding box that inscribes the object is predicted for each of the identified object by point-GNN. We predefine the prototypes based on the prior knowledge about ideal buildings, their energy usage and energy efficiency. For each of the building type $b_i$, we consider a set of prototypes $P_{b_i}$, where $1 \leq i \leq B$ with total $B$ different types of buildings. For each of these prototypes, there are associated predefined specifications of each of their features. To $k$th feature $f_{jk}$ of the prototype $j \in P_{b_i}$ we assign a coefficient that denoted the contribution of the particular feature in deciding the better matching of the identified objects with prototypes.

We calculate the mean $\mu_l, \mu_w, \mu_h$ and standard deviation $\sigma_l, \sigma_w, \sigma_h$ of the length, width and height respectively of the identified bounding boxes for each of the classes. The prototypes are chosen with dimensions $(\mu_l, \mu_w, \mu_h), (\mu_l \pm \sigma_l, \mu_w \pm \sigma_w, \mu_h \pm \sigma_h)$ and $(\mu_l \pm 2\sigma_l, \mu_w \pm 2\sigma_w, \mu_h \pm 2\sigma_h)$ of for all different types of buildings. 

To match the identified buildings to the prototypes, we consider the closest values of the dimensions of the prototypes to that of the buildings. The range for of the five prototypes corresponding to every building type is $(\mu_l - 2.5 \sigma_l, \mu_w - 2.5 \sigma_w, \mu_h - 2.5 \sigma_h) - (\mu_l - 1.5 \sigma_l, \mu_w - 1.5 \sigma_w, \mu_h - 1.5 \sigma_h)$ , $(\mu_l - 1.5 \sigma_l, \mu_w - 1.5 \sigma_w, \mu_h - 1.5 \sigma_h) - (\mu_l - 0.5 \sigma_l, \mu_w - 0.5 \sigma_w, \mu_h - 0.5 \sigma_h)$, 
$(\mu_l - 0.5 \sigma_l, \mu_w - 0.5 \sigma_w, \mu_h - 0.5 \sigma_h) - (\mu_l + 0.5 \sigma_l, \mu_w + 0.5 \sigma_w, \mu_h + 0.5 \sigma_h)$,
$(\mu_l + 0.5 \sigma_l, \mu_w + 0.5 \sigma_w, \mu_h + 0.5 \sigma_h) - (\mu_l + 1.5 \sigma_l, \mu_w + 1.5 \sigma_w, \mu_h + 1.5 \sigma_h)$, and
$(\mu_l + 1.5 \sigma_l, \mu_w + 1.5 \sigma_w, \mu_h + 1.5 \sigma_h) - (\mu_l + 2.5 \sigma_l, \mu_w + 2.5 \sigma_w, \mu_h + 2.5 \sigma_h)$ respectively. 

We can then match each of the identified building to its closest prototype. 
The loss in matching the prototypes of each of the building type is calculated as follows:
\begin{equation}
    \label{eq:loss-matching}
    l_{pro} = \sum_{j=i}^{N} \sum_{i=1}^{F} \phi_i (f_{a}^{ij} - f_{mp}^{ij})
\end{equation}
where, $F$ is the number of features of the buildings, $\phi_i$ is the coefficient corresponding to $i$th features, $f_{a}^{ij}$ is the actual value of the $i$th feature of $j$th example and $f_{mp}^{ij}$ the value of that feature in the matched prototype. $\phi_i$ can be learnt when the data is annotated.

\end{ColorPar}

\begin{ColorPar}{black}
\subsection{Approach 2: Learning with MLP}
\end{ColorPar}
The second part of the system in figure \ref{fig:prototype-pred} classifies the detected objects into classes (prototypes). This network also gives a multi-class probabilities $\{ p_{p_1}, p_{p_2}, \dots, p_{p_T} \}$ corresponding to the $T$ prototypes for each of the detected objects. We compute the cross entropy loss as follows.

\begin{equation}
    l_{pro} = - \frac{1}{N} \sum_{i=1}^N \sum_{j=1}^T z_{p_j}^i log(p_{p_j}^i)
\end{equation}
where $z_{p_j}$ is the one-hot vector indicating the prototype corresponding to the example.

The total loss is given by: 
\begin{equation}
        l_{total} = \alpha l_{cls} + \beta l_{loc} + \gamma l_{pro} + \kappa l_{reg}
\end{equation}
where $\alpha, \beta, \gamma, \kappa$ are the constants to balance losses.

\subsection{Efficiency}

With the use of eBIM-GNN, we get the better and efficient prototypes for the detected objects. To calculate the gain in efficiency, we compare the total energy consumption of the current building with the total energy consumption of the prototype building. The total energy consumption of the detected object is given by $E_o$, and the total energy consumption of suggested prototype is given by $E_p$. We find the percentage gain in total energy efficiency as given in equation \ref{eq:gain}.
\begin{equation}
\label{eq:gain}
    gain_{efficiency} = \frac{E_o - E_p}{E_o} \times 100
\end{equation}


\begin{ColorPar}{black}
\section{Evaluation on Synthetic Data}
\subsection{Dataset}
We consider the KITTI dataset \cite{geiger2012CVPR} that consists of point clouds and camera images. The dataset has 7481 training examples and 7518 testing examples.  Currently due to unavailability of the annotated data consisting of point clouds of buildings, we demonstrate the results on a similar annotated dataset with objects beings cars, pedestrians and cyclists instead of buildings. Since we are considering the KITTI dataset for demonstration, we associate the energy usage to the passenger capacity of each of the three categories. We consider the categories as building types for the sake of convenience. Category 'car' is considered as building type 1, 'pedestrian' as building type 2 and 'cyclist' as building type 3. Once identified, we consider the bounding boxes along with the identified class for to match it with the predefined prototypes.


We get five prototypes corresponding to each class. The ranges of dimensions of bounding boxes for matching the prototypes are as shown in the table \ref{tab:bounding-boxes}. These values are $-0.5 \sigma$ to $+0.5 \sigma$ to each of the dimensions of prototypes. We calculate the total efficiency, as well as the efficiency of each of the building types.

\begin{table*}[ht]
    \caption{Range of dimensions of prototype classes in $(h, w, l)$ format.}
    \label{tab:bounding-boxes}
    \centering
    \begin{tabular}{|c|p{3cm}|p{3cm}|p{3cm}|}
        \hline
        Prototype & Building type 1 & Building type 2 & Building type 3 \\
        \hline
        1 & (1.6971, 1.7051, 4.3538) - (1.8080, 1.7622, 4.6548) & (1.8890, 0.7882, 1.0638) - (1.9667, 0.8725, 1.2325) & (1.8250, 0.6876, 1.9565) - (1.8860, 0.7517, 2.0837)\\
        \hline
        2 & (1.5862, 1.6480, 4.0527) - (1.6971, 1.7051, 4.3538) & (1.8112, 0.7039, 0.8951) - (1.8890, 0.7882, 1.0638) & (1.7640, 0.6235, 1.8293) - (1.8250, 0.6876, 1.9565)\\
        \hline
        3 & (1.4753, 1.5909, 3.7518) - (1.5862, 1.6480, 4.0527) & (1.7334, 0.6196, 0.7264) - (1.8112, 0.7039, 0.8951) & (1.7030, 0.5594, 1.7021) - (1.7640, 0.6235, 1.8293)\\
        \hline
        4 & (1.3644, 1.5338, 3.4508) - (1.4753, 1.5909, 3.7518) & (1.6556, 0.5353, 0.5577) - (1.7334, 0.6196, 0.7264) & (1.6420, 0.4953, 1.5749) - (1.7030, 0.5594, 1.7021)\\
        \hline
        5 & (1.2535, 1.4767, 3.1498) - (1.3644, 1.5338, 3.4508) & (1.5778, 0.4510, 0.3890) - (1.6556, 0.5353, 0.5577) & (1.5810, 0.4312, 1.4477) - (1.6420, 0.4953, 1.5749) \\
        \hline
    \end{tabular}
\end{table*}

We define the energy usage for each of the building types randomly for our experiments. For building type 1 the energy usage is considered as 40 units, for building type 2 it is 20 units, and for building type 3 it is considered as 10 unit of energy. However, each of the prototype has different energy usage values associated with them. These values are as shown in table \ref{tab:ideal-usage}. 

\begin{table}[ht]
    \caption{Predefined energy usage values for each of the building prototypes}
    \label{tab:ideal-usage}
    \centering
    \begin{tabular}{|c|r|r|r|r|}
        \hline
        Prototype & Building type 1 & Building type 2 & Building type 3 \\
        \hline
        1 & 40 & 10 & 20 \\
        \hline
        2 & 38 & 8 & 18 \\
        \hline
        3 & 36 & 6 & 16 \\
        \hline
        4 & 34 & 4 & 14 \\
        \hline
        5 & 32 & 2 & 12  \\
        \hline
    \end{tabular}
\end{table}

\subsection{Results}

\begin{table}[ht]
    \caption{Counts of buildings in the range of dimensions of respective prototype classes }
    \label{tab:counts}
    \centering
    \begin{tabular}{|c|r|r|r|r|}
        \hline
        Prototype & Building type 1 & Building type 2 & Building type 3 \\
        \hline
        1 & 76 & 0 & 0 \\
        \hline
        2 & 273 & 269 & 110 \\
        \hline
        3 & 1451 & 2104 & 145 \\
        \hline
        4 & 886 & 109 & 26 \\
        \hline
        5 & 1 & 21 & 1 \\
        \hline
    \end{tabular}
\end{table}

In order to calculate efficiency after matching with the prototype, we count the number of examples belonging to each of the prototypes. This is shown in table \ref{tab:counts} By multiplying the energy usage corresponding to each of the prototypes of every class and summing it up, we get the total energy usage. We show the energy usage in table \ref{tab:usage} along with the reduction in energy usage corresponding to each of the building type. The total reduction in energy usage is 16.22\%.

\begin{table}[ht]
    \caption{Percentage reduction in energy usage}
    \label{tab:usage}
    \centering
    \begin{tabular}{|c|r|r|r|}
        \hline
        Prototype & Actual usage & Prototype usage & Red. in usage (\%)\\
        \hline
        Building type 1 & 107480 & 95806 & 10.86\\
        \hline
        Building type 2 & 25030  & 15254 & 39.05\\
        \hline
        Building type 3 & 5640   & 4676  & 17.09\\
        \hline
    \end{tabular}
\end{table}

\end{ColorPar}

\section{Conclusion}
\label{sec:conclusion}
eBIM-GNN suggests the energy efficient prototypes for the buildings detected in a 3D point cloud. The building detection and determination of its bounding box is done using the method of Point-GNN. After the determination of bounding box, with the use of other features corresponding to the detected objects, the second part of system determines the appropriate prototype corresponding to the object. We demonstrate our approach on synthetic dataset.

In future, we would like to collect real data of the buildings and tune out the possible scalability issues related with the resolution of the point clouds and further optimize the matching against the prototypes. Our system benefits from the number of prototypes it has in the system to more accurately predict the energy efficiency gains.

\bibliographystyle{plain}
\bibliography{ref}

\end{document}